\documentclass[runningheads]{llncs}
\usepackage{graphicx}
\usepackage{xcolor}
\usepackage{hyperref}
\usepackage{subfiles}
\usepackage{booktabs}
\usepackage{longtable}
\usepackage[normalem]{ulem}
\usepackage{multirow}
\usepackage[noadjust]{cite}

\begin{document}
\title{Brain-Aware Replacements for Supervised
Contrastive Learning in Detection of
Alzheimer’s Disease}
\titlerunning{Brain-Aware Replacements for Supervised Contrastive Learning}

\author{Mehmet Saygın Seyfioğlu\inst{*}\and 
Zixuan Liu \and Pranav Kamath \and Sadjyot Gangolli \and Sheng Wang \and Thomas Grabowski \and Linda Shapiro}
\authorrunning{M. S. Seyfioglu et al.}

\institute{University of Washington, Seattle, WA 98195, USA
\email{\{msaygin,zucksliu,pranavpk,sadjyotg,swang,tgrabow,shapiro\}@uw.edu}}

\maketitle                         
\begin{abstract}

We propose a novel framework for Alzheimer's disease (AD) detection using brain MRIs. The framework starts with a data augmentation method called Brain-Aware Replacements (BAR), which leverages a standard brain parcellation to replace medically-relevant 3D brain regions in an anchor MRI from a randomly picked MRI to create synthetic samples. Ground truth ``hard" labels are also linearly mixed depending on the replacement ratio in order to create ``soft" labels. BAR produces a great variety of realistic-looking synthetic MRIs with higher local variability compared to other mix-based methods, such as CutMix. On top of BAR, we propose using a soft-label-capable supervised contrastive loss, aiming to learn the relative similarity of representations that reflect how mixed are the synthetic MRIs using our soft labels. This way, we do not fully exhaust the entropic capacity of our hard labels, since we only use them to create soft labels and synthetic MRIs through BAR. We show that a model pre-trained using our framework can be further fine-tuned with a cross-entropy loss using the hard labels that were used to create the synthetic samples. We validated the performance of our framework in a binary AD detection task against both from-scratch supervised training and state-of-the-art self-supervised training plus fine-tuning approaches. Then we evaluated BAR's individual performance compared to another mix-based method CutMix by integrating it within our framework. We show that our framework yields superior results in both precision and recall for the AD detection task.

\let\thefootnote\relax\footnotetext{* Corresponding author}

\end{abstract}

\section{Introduction}

Alzheimer's Disease (AD) is an irreversible neurodegenerative condition, which is characterized by atrophy of brain tissue, with distinctive microscopic changes. However, AD-related atrophy is hard to detect, because healthy aging also causes some atrophy. Therefore, for a population-level impact, an abundantly available medical modality, MRI, can be used to detect the disease. Lately, deep-learning-based approaches have become common \cite{liu2020design, zhao2021application}, mostly using the ADNI dataset. However, much of the early work is hard to reproduce due to data-leakage problems \cite{fung2019alzheimer}. Thus further research is needed on the topic.

Contrastive learning has been recently shown to be a powerful technique to learn semantics-preserving visual representations in a self-supervised manner \cite{he2020momentum, chen2020simple}. Based on SimCLR \cite{chen2020simple}, the idea is to create two differently augmented copies (positives) of the anchor image, while considering the rest of the samples within the batch as negatives. Augmentations are a set of parametric transformations, such as random crops, rotations, etc. that aim to preserve semantics of the data while altering them. These positives are then mapped closer in the latent space, while the negatives become further away. This approach is shown to be very effective in natural images \cite{chen2020simple}, as well as in some medical tasks \cite{zhou2019models, tang2021self}. However, the self-supervised contrastive approach has its drawbacks in AD detection, which is a binary classification problem. The assumption that the anchor and the rest of the batch are equally semantically different is incorrect, because it is highly likely that a batch could contain a false negative sample, thus making the training harder.

One way to fix this problem is to use supervised-contrastive learning, which leverages hard labels \cite{khosla2020supervised}. However, this approach has its limitations as using hard labels during pre-training exhausts the entropic capacity of labels, thus leading to sub-optimal fine-tuning performance. Soft labels could be employed during supervised contrastive training, which can be exploited to learn the relative similarity of pairs. CutMix \cite{yun2019cutmix} is a technique known to be very effective in creating soft labels by non-linearly combining images to create synthetic images and labels. A slightly modified version of CutMix has recently been applied in a brain lesion segmentation task \cite{zhang2021carvemix}, where instead of using an arbitrary patch for replacement, lesion-based ROIs are utilized according to the lesion location and geometry. We argue that since AD-related atrophy is widely distributed across different parts of the brain, replacing a big patch, as in \cite{yun2019cutmix}, or focusing on a single ROI, as in \cite{zhang2021carvemix} is not suitable for AD detection, where global understanding of the entire input MRI is essential. Furthermore, for pixel-wise aligned inputs such as ours, replacing a big patch usually creates an easier task for the model, but for pre-training the whole idea is to create difficult tasks so the model will learn more powerful representations.

We propose an augmentation technique for brain MRIs that we call Brain-Aware Replacements (BAR), which utilizes anatomically relevant regions from the Automated Anatomical Labeling Atlas (AAL) for non-linear replacements from a randomly picked MRI into an anchor MRI to produce  synthetic MRIs and soft labels. Compared to CutMix, BAR produces more realistic-looking synthetic MRIs, which leads to higher local variability, thus harder-to-solve synthetic samples. On top of BAR, inspired by \cite{cao2021supervised}, we propose a supervised contrastive pre-training plus fine-tuning framework. However, unlike \cite{cao2021supervised}, our pre-training model aims to learn the relative similarity of representations, reflecting how much the mixed images have the original positives or negatives by optimizing a continuous-value-capable supervised contrastive loss \cite{dufumier2021contrastive}. This way, we do not fully exhaust the entropic capacity of our hard labels, since we only use them to create soft labels and synthetic MRI mixtures through BAR.

Our contributions are two-fold. First, we propose BAR, a novel augmentation strategy that utilizes the AAL to create realistic-looking synthetic samples and soft labels. Second, we show that training a supervised contrastive loss with the soft labels and synthetic MRIs generated through BAR leads to very powerful representation learning. We also show that the pre-trained model can be further fine-tuned utilizing the same labels that were used to create synthetic MRIs and soft labels. To the best of our knowledge, supervised mixture learning with contrastive loss has yet to be investigated, as most of the contrastive mixture learning approaches are conducted in self-supervised fashion \cite{kim2020mixco, kalantidis2020hard}. Also, our work is the first application of supervised contrastive learning within AD classification research. We compare our results with a slightly modified version of CutMix by incorporating it into our framework, as well as state-of-the-art self-supervised and supervised pre-training approaches and show that our approach outperforms them on the AD-vs-cognitively-normal binary classification task. We will share our code at \footnote{https://github.com/aldraus/BrainAwareReplacementsForAD}.

\section{Method}\label{methods}

\textbf{Contrastive Objectives:} The goal of contrastive learning \cite{chen2020simple} is to map samples to a unit hypersphere by preserving semantics, i.e., semantically similar samples are pulled together, and different ones are pushed apart. In the self-supervised approach, an anchor image $X_i$ is augmented twice using a set of transformations, which creates two augmented views, $t_{1}^{i}$ and $t_{2}^{i}$. With these two augmented views, an InfoNCE loss \cite{van2018representation} can be calculated as follows:

\begin{equation}
L_{NCE} = -\sum_{i = 1}^{n}log \frac{{e^{\theta(t_{1}^{i}, t_{2}^{i})}}}{\frac{1}{b}\sum_{j = 1}^{b} e^{\theta(t_{1}^{i}, t_{2}^{j})}}
\end{equation}
where $\theta$ denotes an encoder, $t^j_2|_{j\neq i}$ is a negative sample, $n$ is the number of samples, and $b$ is the number of samples within the batch. For each sample $i$, the model learns to map  $t_{1}^{i}, t_{2}^{i}$ closer while pushing the negatives further away under the assumption that they are equally different from the anchor $t_{1}^{i}$ by optimizing the InfoNCE loss. 

However, for classification problems with a small number of classes such as ours, this approach has its flaws since it is highly probable that $(t_{1}^{i}, t_{2}^{j})_{j\neq i}$ contains false negatives, i.e., the samples that are from the same class as the anchor. To alleviate this problem, a supervised version of InfoNCE could be used as given in \cite{khosla2020supervised}. However, this approach requires the use of all labels during pre-training, which in fact limits the model's performance during the final fine-tuning stage.

\textbf{CutMix Strategy:} Given a set of 3D brain MRIs $X_i|_{i=1}^n$ and their binary annotations $y_i|_{i=1}^n$ stating whether they have AD or not,  it is possible to generate synthetic images $X^p_i|_{i=1}^n$  and soft labels $y^p_i|_{i=1}^n$ by transferring a 3D region from $X_j$ into $X_i$, and modifying the label $y_i$ to be a linear combination of $y_i$ and $y_j$ as follows \cite{yun2019cutmix}:

\begin{equation}
X_i^p =  (1-M) \odot X_i + M \odot X_j
\end{equation}

\begin{equation}
y_i^p = \lambda y_i + (1 - \lambda) * y_j
\end{equation}

Here $M$ denotes a binary mask, and $\lambda$ denotes the pixel-wise combination ratio. This process can be repeated to create a large variety of synthetic images and soft labels and is shown to be effective in natural images. For natural images, local ambiguity generally yields less optimal results, since fine-grained features are mostly localized, thus local connectivity is important. In AD detection however, the atrophy is not localized, but instead generally spread across the whole MRI, so replacing a number of smaller locally disconnected regions instead of a big patch would give more of an insight into the disease. Also, unlike natural images, our MRI data is pixel-wise aligned, thus, estimating the mixture from a big connected region is an easier problem compared to estimating the mixture when a number of smaller patches are replaced. Because the goal is to make pre-training objectives harder, it is more suitable to use a number of smaller patches when replacing parts. This way, the model is implicitly forced to have more of a global understanding.

\textbf{Proposed Framework:} Our framework is based on two ideas: First, to address the problems mentioned in the CutMix section, we propose Brain Aware Replacement (BAR) as an alternative augmentation strategy that non-linearly creates realistic looking mixtures within the dataset by replacing anatomically relevant 3D brain regions. BAR has some advantages over CutMix. Unlike CutMix, the generated images always look realistic, thus there is less distribution shift \cite{gontijo2020tradeoffs}, which in turn helps network training. Also, BAR explicitly forces the model to pay attention to the relationship between medically relevant brain regions, instead of random patches provided by CutMix. Second, to alleviate the problems mentioned in the Contrastive Objectives section, we propose the use of a continuous-valued supervised contrastive objective \cite{dufumier2021contrastive} with soft labels that are produced with BAR. Inspired by \cite{cao2021supervised}, our framework is based on a supervised pre-training plus supervised fine-tuning approach; the overall architecture is shown in Fig 1.

For BAR, given that $\forall i \in [1..N]$, $X_i$ is pixel-wise aligned, in similar fashion to Eq. 2, a 3D binary mask $M$ is generated by sampling regions from the Automated Anatomical Labeling Atlas (AAL) 2 \cite{rolls2015implementation}, which has 62 distinct brain regions when the left and right lobes are merged. The variable $\lambda$ is sampled from a left-skewed beta distribution with $\sigma = 0.2$ and $\beta = 0.8$ and is used for sampling a number of regions from the AAL to create $M$. A number of anatomical brain regions in proportion with $\lambda$ are then carved from a randomly selected sample $X_j$ and are replaced into the same regions of $t_1^i$ based on Eq 2. Eq. 3 is then used to create a pseudo label $y_i^p$ for $t_1^i$ by linearly mixing $y_i$ and $y_j$. This approach helps create a large variety of natural looking inputs, which enables the model to be further fine-tuned using the hard labels that are used to create the synthetic samples through BAR. To prevent the model from focusing on the same regions, $t_{1}^{i}$ is further augmented by Brain Aware Masking (BAM), which fills the $20\%$ of the anatomical brain regions that are left untouched in the swapping stage with random noise. Then, $t_{2}^{i}$ is augmented by either in-painting  \cite{devries2017improved} or out-painting\cite{chen2019self}, and local pixel shuffling \cite{kang2017patchshuffle}, which are all adopted for 3D inputs. During supervised pre-training, $t_{1}^{i}$ and $t_{2}^{i}$ are then used to train a Siamese network. Here a continuous valued supervised contrastive loss is used as given in \cite{dufumier2021contrastive}:

\begin{equation}
   L_{NCE}^c = -\sum_{k=1}^{n}\frac{\varphi(y^p_k, y^p_i)}{\sum_{j=1}^{b} \varphi(y^p_j , y^p_i)} log \frac{{e^{\theta(t_{1}^{i}, t_{2}^{k})}}}{\frac{1}{b}\sum_{j = 1}^{b} e^{\theta(t_{1}^{i}, t_{2}^{j})}}
\end{equation}

\noindent where $\varphi$ denotes a distance kernel between two labels, which in our case are the soft labels of mixtures given in Eq. 3. Hence, we explicitly force our model to learn the relative similarity of augmented versions, and bring similarly mixed MRIs together by focusing on anatomically replaced brain regions. Then a 3D reconstruction (recon) objective \cite{tang2021self} between the anchor MRI and the decoder-processed second augmented copy $t_2^i$ (that does not contain any replacements from another MRI) is trained as follows: 

\begin{equation}
   \mathcal{L}_{recon} = \left \| t_2^i - X_i  \right \|_1
\end{equation}

The final pre-training loss is then calculated as $\alpha_1 \mathcal{L}^c_{NCE} + \alpha_2 \mathcal{L}_{recon}$. We conducted a hyperparameter search based on the validation set, and found that $\alpha_1 = \alpha_2 = 1$ yields the best results.

\begin{figure}[t!]
\centering
\includegraphics[width=0.85\linewidth]{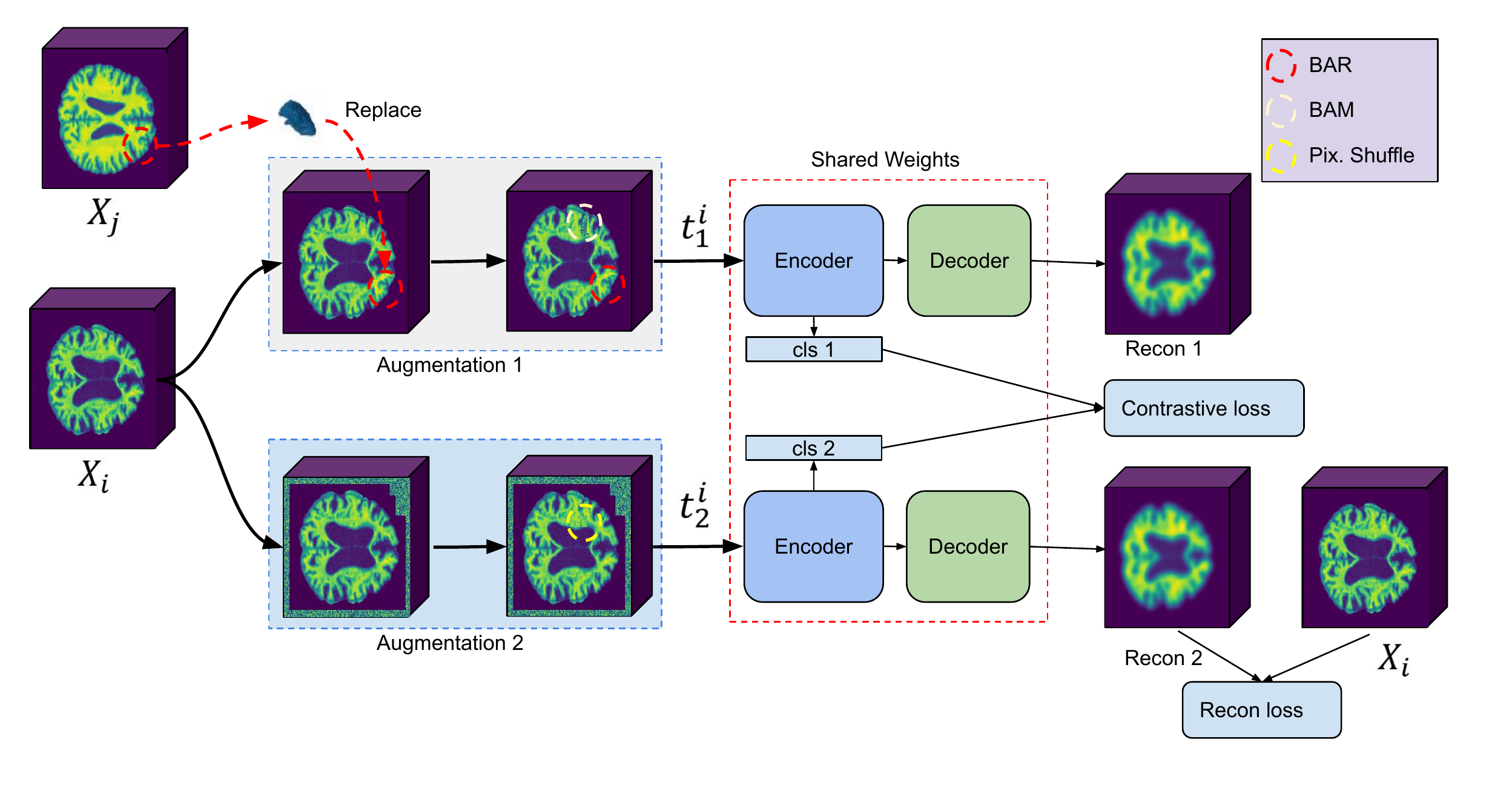}
  \caption{Overall architecture of the proposed method. A number of anatomically-relevant 3D brain regions are taken from $X_j$ and replaced in the same part of $X_i$ where $j \ne i$. The replaced version, $t_1^i$, is followed by BAM, where another brain region is replaced with random noise. Then $t_2^i$ undergoes inner-outer cuts and pixel shuffling, and a Siamese ViT is used to get the cls tokens for both views, which in turn are used to compute the contrastive loss. Furthermore, a decoder is used to generate an image from $t_2^i$ (which has no replacements) and a reconstruction loss is calculated between the recon 2 and the anchor $X_i$.}
  \label{fig:network_temp}
\end{figure}

\textbf{Model Architecture}
We utilize a 3D Vision Transformer (ViT) \cite{hatamizadeh2022unetr} as our encoder with 10 layers and 12 attention heads. Our ViT takes a 3D input volume with resolution $(H,W,D)$ where $H,W,D$ are each 96, and sequences it in non-overlapping flattened patches with resolutions of $16 \times 16 \times 16$. This creates $H \times W \times D/16^3$ = 216 patches for each MRI. All patches are projected into a 768-dimensional embedding space and added on the learnable positional embeddings. The learnable $[cls]$ token is added at the beginning of the sequence of embeddings, so each MRI is represented with a $217 \times 768$ dimensional matrix. Then we use multi-head self-attention and multilayer perceptron sublayers, both of which utilize layer normalization \cite{ba2016layer}. For our decoder, we use 2 layers of Convolutional Transpose layers which reconstruct the MRIs from $216 \times 768$ latent representations ($[cls]$ is not used during reconstruction). The model outputs two tensors, a reconstructed output for the recon objective and the cls token for the contrastive objective.

\section{Experimental Settings}

\textbf{Data:} We used the Alzheimer's Disease Neuroimaging Initiative (ADNI) dataset\footnote{adni.loni.usc.edu} in this work. The ADNI was launched in 2003 as a public-private
partnership, led by Principal Investigator Michael W. Weiner, MD. The primary goal of ADNI has been to
test whether serial MRI, positron emission tomography, other
biological markers, and clinical and neuropsychological assessment can be combined to measure the
progression of mild cognitive impairment and early AD. We include all participants across the ADNI 1-2-3 cohorts that have structural T1 MRI scans divided in 246 Alzheimer's patients (AD) and 597 healthy controls (CN), which totals to 3306 MRIs. The data is first registered to the ICBM template, then skull stripping and bias field correction are conducted, and the resulting MRIs are resampled to 96x96x96 voxels along the sagittal, coronal and axial dimensions. The data is split into training and testing sets that do not have overlapping subjects, which prevents data leakage \cite{fung2019alzheimer}. The average age of subjects in the training and testing sets are roughly similar and around $77$. 5-fold cross validation is conducted, and every time one of the folds is used to select the model parameters and tested on the individual test set.

\textbf{Implementation Details:} We used Pytorch \cite{paszke2019pytorch} and MONAI \footnote{https://monai.io/} to implement our models. In all our experiments, we used the ADAM optimizer, with a learning rate of $10^{-4}$ for pre training and $3*10^{-5}$ for fine tuning stages. $\sigma = 0.18$ is used for our RBF kernel $\varphi$. We trained our models using 4 NVIDIA RTX A4000 GPUs, having 16Gb VRAM each; a batch size of 4 is employed during pre-training due to computation limits of 3D modalities and a batch size of 12 is used during fine tuning. For fine tuning, we used 2 layers of MLP with 128 and 64 neurons that we attach on top of our $[cls]$ token and trained with binary cross entropy loss. For augmentations, we used MONAI's RandCoarseDropout and RandCoarseShuffle functions with holes = 6, spatial size = 5 for inner cut, holes = 6, spatial size = 20 for outer cut, and holes = 10 and spatial size = 5 for pixel shuffling.

\textbf{Experimental Design:} We compared the performance of our proposed framework against: 1) Training a model from scratch, 2) Self-Supervised pre-training + fine-tuning, 3) Modified CutMix based supervised pre-training + fine tuning. For training from scratch, we used a 3D ViT as our encoder. For the self-supervised approach, we tested with three different settings to see the individual contributions of contrastive objective, reconstruction objective and their combination. For the modified Cutmix, we replace a number of smaller 3D patches from the target MRI to the anchor MRI, instead of a single patch. For augmentations, we used inner and outer cutouts with equal probability for both augmented views, followed by pixel shuffling. Finally, we tested the performance of BAR against CutMix.

\section{Experimental Results}

The results for the AD vs NC task are shown in Table 1. As expected, both self-supervised and supervised pre-training outperform training from scratch. For the self-supervised approach, when trained alone, the contrastive objective yields the worst results, especially on recall. We hypothesize that this is caused by the large number of false positives due to the binary nature of the problem. Because the CN case is more abundant in the training data, that class is more affected, thus explaining the poor recall rate. In some cases, the anchor might even be pushed apart against a sample from the same subject, which is non-optimal. Interestingly, when combined with the recon objective, the contrastive objective provides a slight boost to the performance. Recon stabilizes the learning when combined with the contrastive loss (which is tricky to train as it depends on the intensity of the masking ratio from inner-outer cuts in earlier iterations, and it is quite unstable) and grants a performance boost. Finally, we compare CutMix with BAR, and see that BAR outperforms CutMix both with and without the recon objective. BAR is especially better in precision, which shows that it is better in detecting AD related atrophy. Also, in both cases, using the recon loss during pre-training yields a substantial performance boost.

\begin{table}[]

\begin{tabular}{@{}l|l|l|l|l@{}}
\toprule
Framework                                                                                               & Method     & \multicolumn{1}{c|}{Precision} & \multicolumn{1}{c|}{Recall} & \multicolumn{1}{c}{Accuracy} \\ \midrule
No Pre Training                                                                                                  & ViT from scratch    & $74.38\pm7$                               & $85.6\pm3.1$                           & $80.83\pm3$                             \\ \midrule
\multirow{3}{*}{\begin{tabular}[c]{@{}l@{}}Self Supervised Pre-Training \\           + Fine Tuning\end{tabular}} & Contrastive         & $78.42\pm4.5$                             & $81.18\pm1.6$                          & $80.1\pm1.9$                            \\
                                                                                                                 & Recon               & $78.6\pm5$                                & $85.57\pm1.1$                          & $82.69\pm2.5$                           \\
                                                                                                                 & Contrastive + Recon & $80.2\pm4.1$                              & $85.77\pm2$                            & $83.4\pm1.7$                            \\ \midrule
\multirow{4}{*}{\begin{tabular}[c]{@{}l@{}}Supervised Pre-Training\\           +Fine Tuning\end{tabular}}        & CutMIX              & $83.06\pm4.8$                             & $87.08\pm3.5$                        & $85.29\pm2.8$                           \\
                                                                                                                 & CutMIX + Recon      & $84.6\pm3.8$                              & $87.9\pm2.2$                           & $86.4\pm1$                              \\
                                                                                                                 & BAR                 & $84.7\pm3.3$                              & $87.6\pm2.1$                           & $86.3\pm1.1$                            \\
                                                                                                                 & \textbf{BAR + Recon}         & $\textbf{86.24}\pm\textbf{3}$                               & $\textbf{88.08}\pm\textbf{2.3}$                          & $\textbf{87.22}\pm\textbf{0.8}$                           \\ \bottomrule
\end{tabular}
\linebreak
\caption {Fine-tuning results for AD vs. CN case; best is shown in \textbf{bold} } \label{tab:results} 
\end{table}

\textbf{Ablation Study on Brain Aware Masking in Self-Supervised Case:} We test the performance of BAM, (i.e., we randomly selected and filled 3D anatomical brain regions with noise) against the use of inner and outer cuts in a self-supervised manner. When fine-tuned, the performance is comparable to inner outer cuts with an overall accuracy of $83.54 \pm 1.8$ when trained with Contrastive + Recon with a similar drop ratio used in inner-outer cuts.

\textbf{Ablation Study on the Selection of Beta Distribution for BAR:} We tried two different beta distributions for sampling brain regions in BAR, a left skewed one with parameters of beta(0.2, 0.8) and a uniform distribution with beta(1,1). We obtained an overall accuracy of $86.9 \pm 1.5$ with the left skewed one as opposed to $87.2 \pm 1.3$ with the uniform one. We argue that the replacement ratio sampled from the left skewed beta distribution makes somewhat of an easier objective with less replacements and thus is easier to solve. However, more research is needed to find the optimal replacement ratio.

\textbf{Ablation Study on Further Transferability:} To see how much further transferability is possible, we froze the ViT encoder in the BAR framework and trained an MLP with the cls tokens of the encoder. We obtained an accuracy of $85.2$ which shows that there is further room for the same features to be used for fine tuning, as the fine-tuned model yields about $87.22$. We argue that this is the case because we do not directly use our hard labels during pre-training but use them for creating soft labels and realistic looking synthetic images instead, thus their entropic capacity is not fully exhausted during the pre-training phase.

\textbf{Ablation Study on Directly Using the Hard-labels During Pre-training:} We also compared our soft-label supervised contrastive learning + fine-tuning approach against hard-label supervised contrastive learning + fine-tuning approach. To that end, we utilized hard-labels and no replacements during pre-training of supervised contrastive loss \cite{khosla2020supervised} + recon loss, using inner outer cuts and pixel shuffling for both $t_1^i$ and $t_2^i$. This approach is produces lower quality embeddings compared to the soft-label approach as it yields an accuracy around $83.7\%$ by training an MLP on top of the frozen encoder, and its fine tuning results are $84.7\%$.

\section{Discussion and Conclusion}
We proposed a new framework for AD detection that combines a novel augmentation strategy, BAR, which leverages 3D anatomical brain regions to create synthetic MRIs and labels. We showed that, when pre-trained with the synthetic samples, a continuous valued supervised contrastive loss is very effective for the AD detection task. We experimented on the public dataset ADNI and showed that our approach outperforms training from scratch as well as self-supervised approaches. Furthermore, we compared BAR with (CutMix), a popular synthetic data generation strategy into our framework, and showed that, for the AD detection task, using medically relevant brain regions is superior to replacement with arbitrary patches. For future work, We plan to expand our dataset to see how scaleable this framework is with larger datasets.

\subsubsection*{Acknowledgements.}Data used in preparation of this article were obtained from the Alzheimer’s Disease Neuroimaging Initiative
(ADNI) database. As such, the investigators within the ADNI contributed to the design
and implementation of ADNI and/or provided data but did not participate in analysis or writing of this report.
A complete listing of ADNI investigators can be found at http://adni.loni.usc.edu/wp-content/uploads/how\_to\_apply/ADNI \_Acknowledgement\_List.pdf


\bibliographystyle{unsrt}

\bibliography{references} 

\newpage

\section{Supplementary Material}

\textbf{1. Visual Comparison of CutMix and BAR}
Synthetic MRIs generated by CutMix and BAR is shown in Fig 2.
\begin{figure}[h!]
\centering
\includegraphics[width=0.85\linewidth]{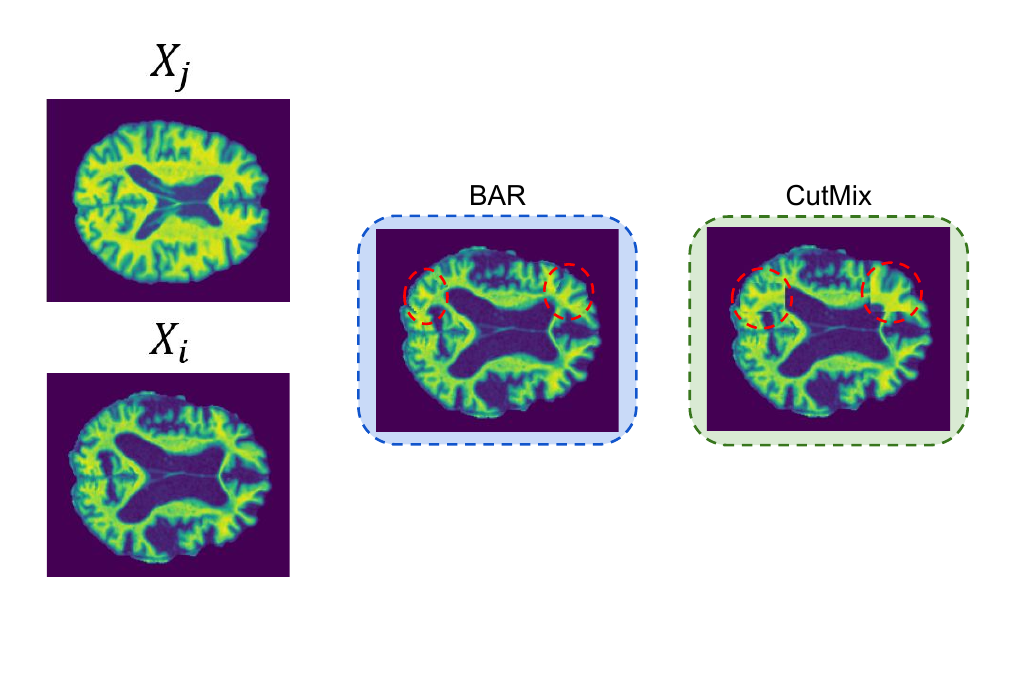}
  \caption{Comparison between a sample generated with BAR and CutMix given an anchor $X_i$ and a random MRI $X_j$ on axial view. For BAR, two random regions are selected from the AAL atlas (\textit{Superior frontal gyrus, medial orbital} and \textit{Superior frontal gyrus, dorsolateral}). These regions are taken from $X_j$ and replaced in the same parts of $X_i$. For CutMix, square-regions are selected from similar regions on $X_j$ and replaced in $X_i$ in the same fashion (marked by red circles). Notice how BAR produces more realistic-looking synthetic MRIs as random patches often are too bulky and cutting/replacing regions from lateral ventricle.}
  \label{fig:BARVSCUTMIX}
\end{figure}

\noindent\textbf{2. Attention Visualization of CutMix and BAR}
We used Attention Rollout \cite{dosovitskiy2020image}, which yields averaged attention weights across all layers and heads. We analysed two cases, an AD sample, and a CN sample. The average attention outputs are shown in Fig. 3 and Fig. 4, respectively for AD and CN (40th-70th slices are shown with increments of 10 for all views.). In both cases CutMix based model erroneously classified the given sample and BAR based model made a correct prediction. BAR attends regions more globally (Fig. 3) for the AD case and also correctly chooses not to focus on non-AD atrophy (Fig. 4).

\begin{figure}[h!]
\centering
\includegraphics[width=1.05\linewidth]{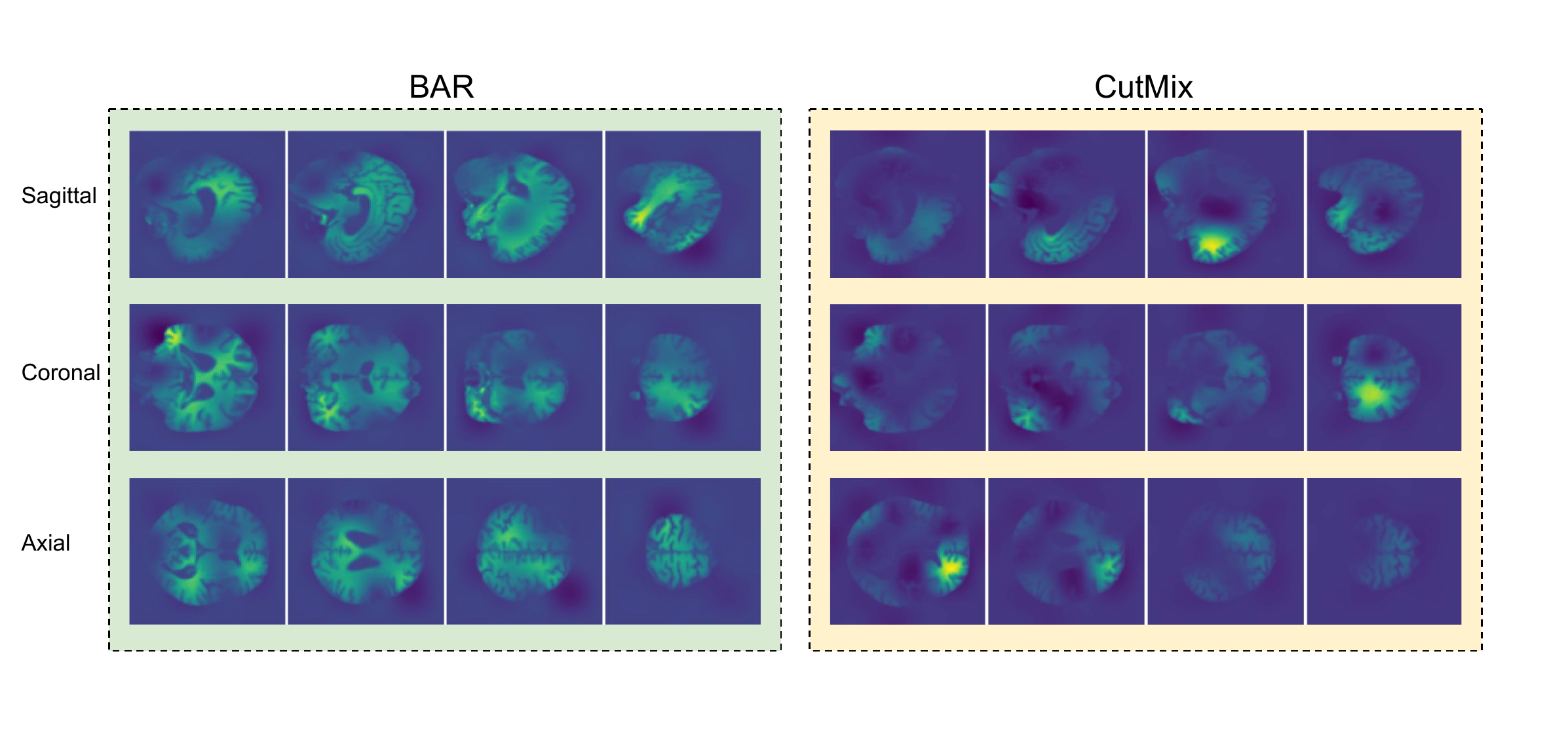}
  \caption{Attention Rollout results for the AD case, slices between 40th-70th are shown with increments of 10 for all views.}
  \label{fig:BARVSCUTMIXAD}
\end{figure}

\begin{figure}[h!]
\centering
\includegraphics[width=1.05\linewidth]{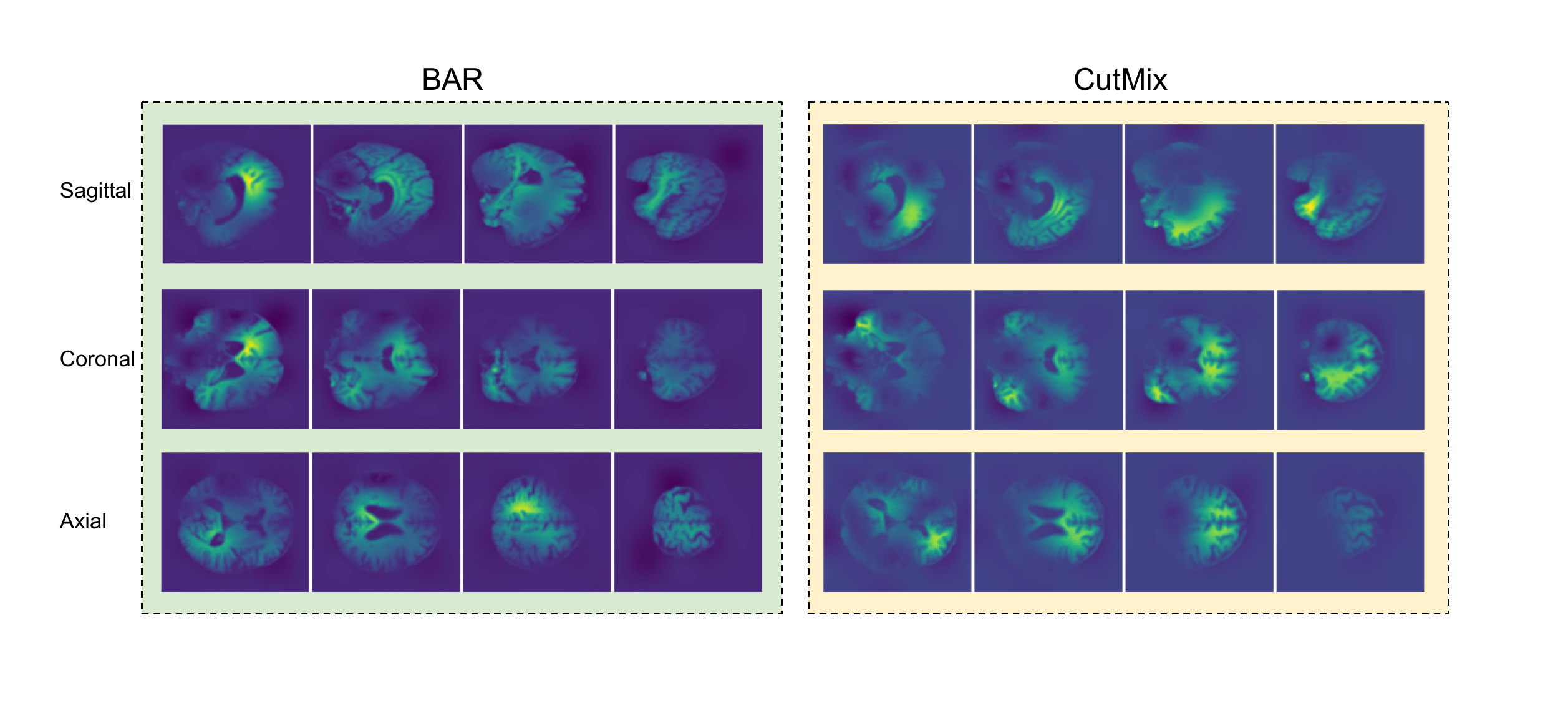}
  \caption{Attention Rollout results for the CN case, slices between 40th-70th are shown with increments of 10 for all views.}
  \label{fig:BARVSCUTMIXCN}
\end{figure}

\end{document}